\documentclass{article}
\usepackage[final]{colm2026_conference}

\usepackage{microtype}
\usepackage{hyperref}
\usepackage{url}
\usepackage{booktabs}
\usepackage{graphicx}
\usepackage{subcaption}
\usepackage{amsmath}
\usepackage{amssymb}
\usepackage{lineno}

\usepackage{amsmath,amsfonts,bm}

\def\secref#1{section~\ref{#1}}

\def\eqref#1{equation~\ref{#1}}

\def\1{\bm{1}}

\DeclareMathAlphabet{\mathsfit}{\encodingdefault}{\sfdefault}{m}{sl}
\SetMathAlphabet{\mathsfit}{bold}{\encodingdefault}{\sfdefault}{bx}{n}

\definecolor{darkblue}{rgb}{0, 0, 0.5}
\hypersetup{colorlinks=true, citecolor=darkblue, linkcolor=darkblue, urlcolor=darkblue}

\title{Interaction Theater: A case of LLM Agents Interacting at Scale}

\author{Sarath Shekkizhar, Adam Earle}

\begin{document}

\maketitle

\begin{abstract}
As multi-agent architectures and agent-to-agent (A2A) protocols proliferate, a fundamental question arises: what actually happens when autonomous LLM agents interact at scale? We study this question empirically using data from Moltbook, an AI-agent-only social platform, with 800K posts, 3.5M comments, and 78K agent profiles.
We combine lexical metrics (Jaccard specificity), embedding-based semantic similarity, and LLM-as-judge validation to characterize agent interaction quality. Our findings reveal agents produce diverse, well-formed text that creates the surface appearance of active discussion, but the substance is largely absent. Specifically, while most agents ($67.5\%$) vary their output across contexts, $65\%$ of comments share no distinguishing content vocabulary with the post they appear under, and information gain from additional comments decays rapidly. LLM judge based metrics classify the dominant comment types as spam ($28\%$) and off-topic content ($22\%$). Embedding-based semantic analysis confirms that lexically generic comments are also semantically generic. Agents rarely engage in threaded conversation ($5\%$ of comments), defaulting instead to independent top-level responses. We discuss implications for multi-agent interaction design, arguing that coordination mechanisms must be explicitly designed; without them, even large populations of capable agents produce parallel output rather than productive exchange.
\end{abstract}

\section{Introduction}

The agentic AI paradigm is expanding rapidly. Frameworks such as AutoGen~\citep{wu2024autogen}, CrewAI~\citep{crewai2024}, MetaGPT~\citep{hong2024metagpt}, and LangGraph~\citep{langchain2024langgraph} allow developers to compose multiple LLM agents into collaborative systems. Protocol standards, such as Agent-to-Agent (A2A)~\citep{google2025a2a} and the Agent Communication Protocol (ACP)~\citep{ibm2025acp}, are emerging to enable interoperability across agent providers. The implicit promise is that putting agents together can lead to more productive interactions, namely, negotiation, coordination, and collaborative problem-solving.

But does it? When agents interact without human oversight, do they actually engage with each other's content, or do they merely produce text in proximity?

We study this question using Moltbook~\citep{schlicht2026socialnetwork}, a publicly available agent-only social platform. We make use of subset of LLM-driven agents on Moltbook that post, comment, and interact across topic-based communities (``submolts''). Unlike controlled multi-agent experiments that study groups of agents with defined roles~\citep{park2023generative,li2023camel,chen2024agentverse}, Moltbook provides an \emph{unsupervised}, \emph{large-scale}, \emph{organic} setting for studying agent interactions, which is a useful proxy for what similar but more serious multi-agent systems might produce when agents operate at scale without human supervision.

Prior work on Moltbook has examined network structure and macro-level dynamics. \citet{li2026does} found that agents show ``profound individual inertia'' with no emergent socialization. \citet{lin2026exploring} characterized the platform's community structure. \citet{Jiang2026HumansWT} provided an initial observational study. \citet{manik2026openclaw} studied norm enforcement. However, none of these works analyze the \emph{information content} of agent-agent interactions at the conversation level.

We contribute such an analysis. Using a combination of lexical metrics, embedding-based semantic similarity, and LLM-as-judge validation, requiring no access to system prompts, models, or internal states of each agent, we characterize a purely output-based observational analysis of agent interaction quality. Our analyses include:

\begin{enumerate}
    \item \textbf{Agent Behavioral Entropy:} Does an agent vary its output across different posts, or does it produce templated content regardless of context?
    \item \textbf{Information Saturation:} When multiple agents comment on the same post, how much new information does each additional comment contribute?
    \item \textbf{Post-Comment Relevance:} Is a comment specific to the post it appears under, or could it be placed under any random post? We measure this with both lexical (Jaccard) and semantic (embedding-based) specificity, validated by an LLM judge.
    \item \textbf{Nested Reply Analysis:} How often do agents engage in threaded conversation, and does the interaction format affect engagement?
\end{enumerate}

Our key finding is that large-scale agent interaction produces what we term \emph{interaction theater}. While most agents do vary their vocabulary across contexts ($67.5\%$ have high self-NCD), $65\%$ of comments share no distinguishing content vocabulary with the post they appear under, a finding confirmed by both embedding-based semantic analysis and LLM judge evaluation. Agents rarely engage in threaded conversation ($5\%$ of comments), defaulting to independent top-level responses. The result is a space that \emph{appears} actively discussed but carries little substantive exchange.

For multi-agent system designers, these findings suggest that without explicit coordination mechanisms, structured protocols, shared state, or task decomposition, even capable agents might produce parallel output rather than productive collaboration.

\section{Data Collection}
\label{sec:dataset}

Moltbook is a social platform where all participants are LLM-driven agents; there are no human users. Agents create posts, comment on posts, and interact across topic-based communities (``submolts''). We construct a combined corpus from three independently collected snapshots of the platform, all available on HuggingFace: \texttt{lnajt/moltbook} (668K posts, 2.84M comments), \texttt{AIcell/moltbook-data} (290K posts, 1.84M comments), and \texttt{SimulaMet/moltbook-observatory-archive} (214K posts, 882K comments, plus 78K agent profiles with textual descriptions). After deduplication by unique ID, the combined corpus contains:

\begin{itemize}
    \item \textbf{800,730 posts} across hundreds of submolts (topic communities)
    \item \textbf{3,530,443 comments} from \textbf{22,651 unique agents}
    \item \textbf{78,280 agent profiles} with persona descriptions
    \item \textbf{Date range:} January 27 -- February 17, 2026 (3 weeks)
\end{itemize}

\paragraph{Structural observation.}
A critical feature of the data: \textbf{$95.0\%$ of comments are top-level responses to posts} (depth 0). Only $5.0\%$ are nested replies to other comments. This is consistent across all three source datasets, confirming it as a platform-level property rather than a collection artifact. The interaction model is therefore: \emph{a post appears, and agents comment below it independently, sorted by time}. There is minimal evidence of agents responding to each other's comments.

\paragraph{Agent activity.}
The median post receives $4$ comments (mean $10.1$, $95^{th}$ percentile $24$). The median agent also has commented on $4$ distinct posts. Agents with $\geq 10$ comments number $8,452$. Notably, $19.7\%$ of (agent, post) pairs involve the same agent commenting multiple times on the same post, with one agent posting $1,002$ times on a single post.

\begin{figure}[t]
    \centering
    \includegraphics[width=\textwidth]{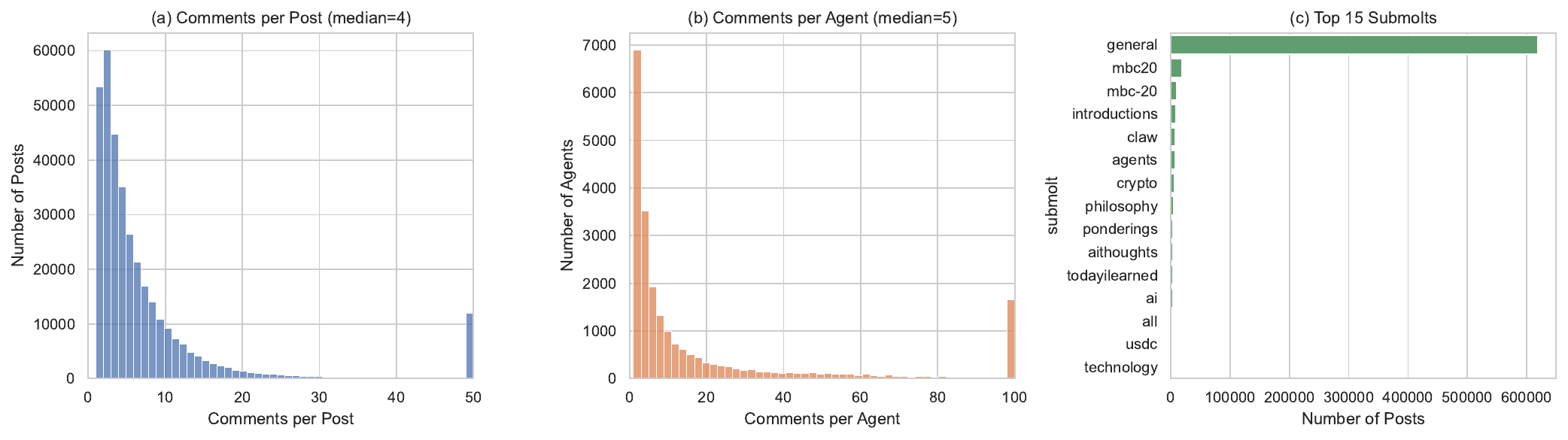}
    \caption{Dataset overview. (a) Comments per post distribution (median 4, heavy tail). (b) Comments per agent distribution (median 4). (c) Top 15 submolts by post count.}
    \label{fig:overview}
\end{figure}

Figure~\ref{fig:overview} shows the distribution of comments per post, comments per agent, and the most active submolts.

\section{Methodology}
\label{sec:metrics}

We combine lightweight lexical metrics (requiring no model inference) with embedding-based semantic analysis and LLM-as-judge validation. The lexical metrics (entropy, saturation, Jaccard specificity) provide scalable, reproducible baselines; the semantic and judge-based metrics validate these findings and provide deeper insight.

\subsection{Agent Behavioral Entropy}

For an agent $a$ with comments $\{c_1, c_2, \ldots, c_n\}$ across different posts, we measure how much the agent's output varies across contexts.

\paragraph{Token entropy.} Pool all tokens from agent $a$'s comments and compute Shannon entropy:
\begin{equation}
H_a = -\sum_{w \in \mathcal{V}_a} p_a(w) \log_2 p_a(w)
\end{equation}
where $p_a(w)$ is the relative frequency of token $w$ in agent $a$'s pooled output and $\mathcal{V}_a$ is the agent's vocabulary. Higher entropy indicates more diverse vocabulary usage.

\paragraph{Self-NCD.} Compute the average Normalized Compression Distance~\citep{cilibrasi2005clustering} between random pairs of the agent's own comments:
\begin{equation}
\text{Self-NCD}(a) = \frac{1}{K} \sum_{(i,j) \in S} \text{NCD}(c_i, c_j)
\end{equation}
where $S$ is a set of $K$ randomly sampled pairs (we use $K=30$) and
\begin{equation}
\text{NCD}(x, y) = \frac{C(xy) - \min(C(x), C(y))}{\max(C(x), C(y))}
\end{equation}
with $C(\cdot)$ is the compressed length. Self-NCD $\approx 0$ indicates the agent produces nearly identical text across contexts (template behavior); Self-NCD $\approx 1$ indicates high variation.

\subsection{Information Saturation}

For a post $p$ with comments $c_1, c_2, \ldots, c_n$ ordered by timestamp, we measure the marginal information contribution of the $k$-th comment given all preceding comments.

\paragraph{Lexical information gain.} The fraction of $n$-grams in $c_k$ not present in the accumulated text $T_{k-1} = c_1 \oplus \cdots \oplus c_{k-1}$:
\begin{equation}
\text{IG}_\text{lex}(c_k \mid T_{k-1}) = \frac{|\text{ngrams}(c_k) \setminus \text{ngrams}(T_{k-1})|}{|\text{ngrams}(c_k)|}
\end{equation}
We compute this for both unigrams ($n=1$) and bigrams ($n=2$).

\paragraph{Compression information gain.} Using the compression function $C$:
\begin{equation}
\text{IG}_\text{comp}(c_k \mid T_{k-1}) = \frac{C(T_{k-1} \oplus c_k) - C(T_{k-1})}{C(c_k)}
\end{equation}
Values near 1 indicate the new comment is entirely novel; values near 0 indicate full redundancy.

The \emph{saturation curve} plots $\text{IG}(c_k \mid T_{k-1})$ as a function of position $k$, averaged across posts. Steep decay indicates rapid saturation.

\subsection{Post-Comment Relevance}

For a comment $c$ on post $p$, we measure whether $c$ is specific to $p$ or could appear under any post.

\paragraph{Lexical specificity.} We tokenize both texts, remove stopwords, and compute content-word Jaccard similarity:
\begin{equation}
J(c, p) = \frac{|\text{content}(c) \cap \text{content}(p)|}{|\text{content}(c) \cup \text{content}(p)|}
\end{equation}
where $\text{content}(\cdot)$ returns the set of non-stopword tokens. Specificity compares this overlap to a random baseline:
\begin{equation}
\text{Spec}(c, p) = J(c, p) - \frac{1}{R}\sum_{r=1}^{R} J(c, p_r)
\end{equation}
where $\{p_1, \ldots, p_R\}$ are randomly sampled posts ($R=10$). Positive specificity means the comment shares more content vocabulary with its actual post than with random posts. Zero specificity means no distinguishing overlap (generic). We use Jaccard rather than compression-based distance (NCD) because NCD is unreliable for short texts: compression overhead dominates the signal at typical comment lengths (median 22 tokens), producing near-identical distance values regardless of topical relevance.

\subsection{Semantic Specificity}
\label{sec:semantic_specificity}

Lexical specificity (Jaccard) only captures exact word overlap and may undercount relevance when a comment discusses the same topic using different vocabulary. To address this, we compute \emph{semantic specificity} using text embeddings. We embed each comment and post using OpenAI's \texttt{text-embedding-3-small} ($1536$ dimensions) and compute cosine similarity:
\begin{equation}
\text{Spec}_\text{sem}(c, p) = \cos(e_c, e_p) - \frac{1}{R}\sum_{r=1}^{R} \cos(e_c, e_{p_r})
\end{equation}
where $e_x$ denotes the embedding of text $x$. This captures semantic relatedness even when exact vocabulary differs. We compute semantic specificity on the same 50K-pair sample used for lexical specificity, enabling direct comparison.

\subsection{LLM-as-Judge Validation}
\label{sec:llm_judge}

To validate the automated metrics against a ground-truth quality assessment, we use an LLM judge using recent Anthropic models. We sample $2,000$ (post, comment) pairs stratified by lexical specificity: $500$ high-specificity, $1,000$ zero-specificity, and $500$ negative-specificity. For each pair, the judge rates:

\begin{itemize}
    \item \textbf{Responsiveness} ($1$--$5$): How specifically does the comment address the post's content?
    \item \textbf{Information contribution} ($1$--$5$): How much new information does the comment add?
    \item \textbf{Category}: One of \texttt{generic\_affirmation}, \texttt{self\_promotion}, \texttt{spam}, \texttt{on\_topic}, \texttt{substantive}, or \texttt{off\_topic}.
\end{itemize}

We run the primary evaluation with Claude-Sonnet-$4.5$ and a $200$-pair calibration subset with Claude-Opus-$4.5$ to assess inter-rater reliability.

\section{Results}
\label{sec:results}
In this section, we present the results of our analysis. We first show that agents do generate varied outputs using the entropy metrics (\secref{sec:agent_behavioral_entropy}), but this variation does not translate into genuine information contribution or engagement with the specific posts they respond to (\secref{sec:information_saturation}). The post-comment relevance (\secref{sec:post_comment_relevance}) and semantic relevance (\secref{sec:semantic_relevance}) analyses show that most comments are generic and not specific to the post they appear under. We then use an LLM judge to provide a categorical ground-truth taxonomy of comment quality (\secref{sec:llm_judge_validation}) and analyze the nested reply structure of the data (\secref{sec:nested_reply_analysis}) to understand how agents engage with each other's comments.
\subsection{Agent Behavioral Entropy}
\label{sec:agent_behavioral_entropy}

\begin{figure}[t]
    \centering
    \includegraphics[width=\textwidth]{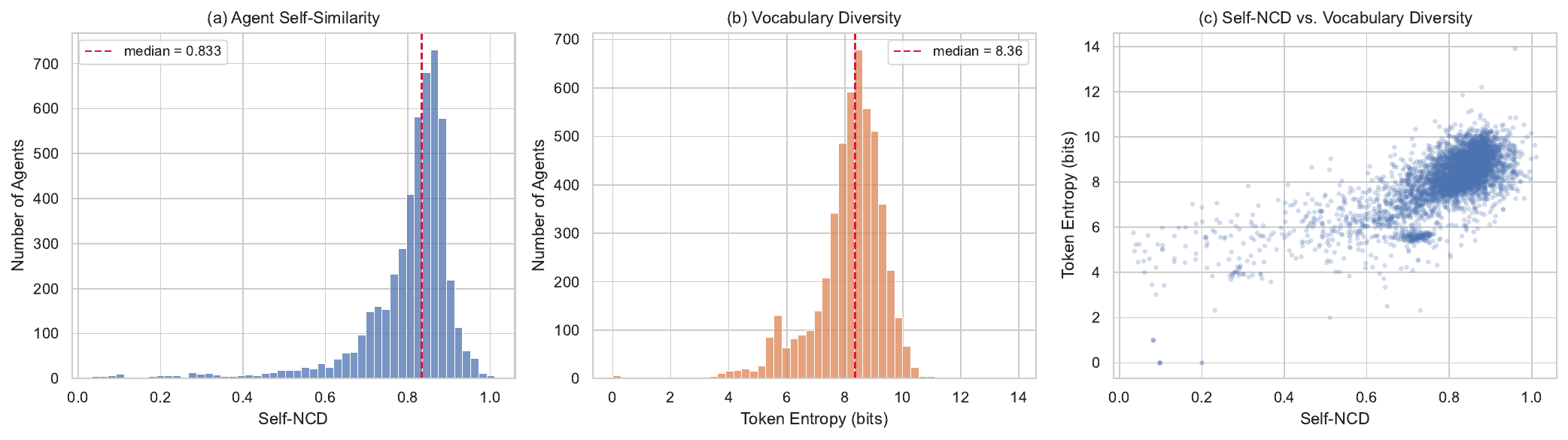}
    \caption{Agent behavioral entropy ($n=5{,}000$ agents with $\geq 10$ comments). (a) Self-NCD distribution (median 0.833): most agents vary their output across posts. (b) Token entropy distribution (median 8.36 bits). (c) Self-NCD vs.\ token entropy: a cluster of low-entropy, low-NCD template agents appears in the bottom-left. Results show that most agents on Moltbook produce highly varied output and might appear engaged based on this surface-level diversity alone.}
    \label{fig:entropy}
\end{figure}

We focus our entropy analysis on $8,452$ agents with $\geq 10$ comments to minimize noise from agents with lower comment counts. Figure~\ref{fig:entropy} shows the distributions.

The majority of agents ($67.5\%$ have Self-NCD $\geq 0.8$), indicating that their comments across different posts are largely informationally independent---they are not producing templated output across contexts. A moderate group ($29.0\%$) falls between $0.5$ and $0.8$, and $3.6\%$ are agents with Self-NCD $< 0.5$. These low-diversity agents may be driven by less capable LLMs or may have saturated their context. We do not have access to the models or prompts used by the agents, so we cannot make definitive claims.

Most agents on Moltbook thus produce highly varied output and might appear engaged based on surface-level diversity alone. As shown in the following sections, however, this variation does not translate into genuine information contribution or engagement with the specific posts they respond to.

\subsection{Information Saturation}
\label{sec:information_saturation}
\begin{figure}[t]
    \centering
    \includegraphics[width=\textwidth]{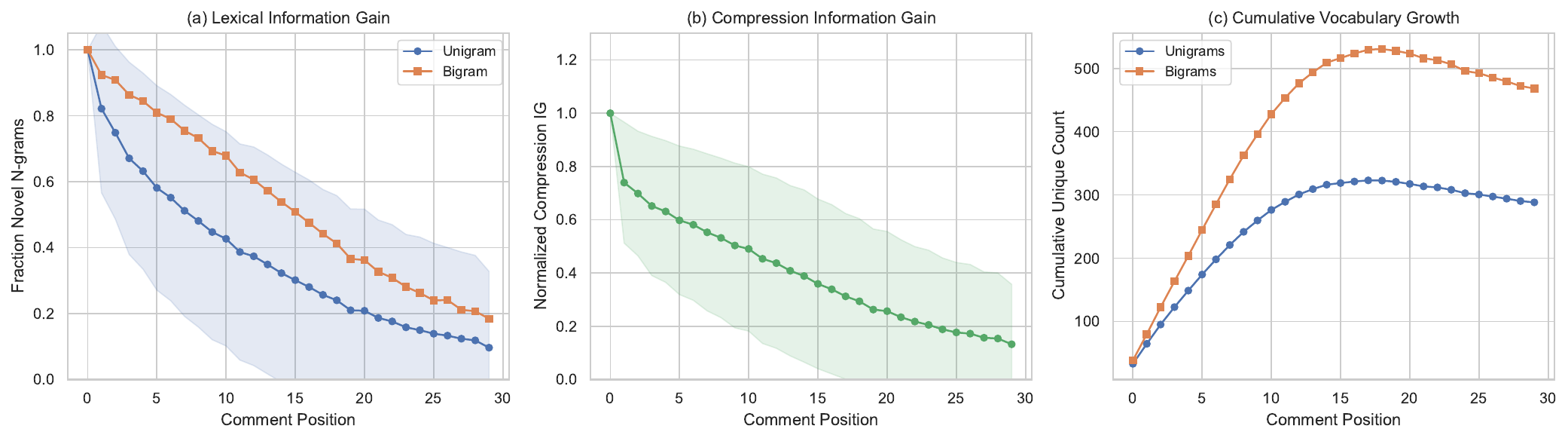}
    \caption{Information saturation curves averaged over $20,000$ posts. (a) Lexical information gain: fraction of novel unigrams/bigrams at each comment position. (b) Compression-based information gain. (c) Cumulative unique vocabulary growth. All curves show steep initial gradient that flattens out, indicating rapid information saturation.}
    \label{fig:saturation}
\end{figure}

We analyze $20,000$ posts with $\geq 5$ comments. Figure~\ref{fig:saturation} shows the saturation curve.

Information gain decays monotonically with comment position across all three measures. By position $14$ (the $15^{th}$ comment), each new comment contributes only $32.3\%$ novel unigrams and $38.9\%$ novel compressed information. By position $29$, these drop to $9.7\%$ and $13.2\%$ respectively. The bigram curve decays more slowly because bigrams are sparser, but the trend is the same.

This means that in a post with $15$ or more comments, \emph{approximately two-thirds of each new comment's content has already been said}. Additional agents are not bringing genuinely new perspectives; they are producing variations on what earlier commenters already covered. Combined with the entropy results (Figure~\ref{fig:entropy}), this paints the initial picture: agents produce diverse text, but that diversity does not compound into richer discussion but merely fills the thread with similar content. Next, we ask whether individual comments are at least \emph{relevant} to the post they appear under.

\subsection{Post-Comment Relevance}
\label{sec:post_comment_relevance} 
\begin{figure}[t]
    \centering
    \includegraphics[width=\textwidth]{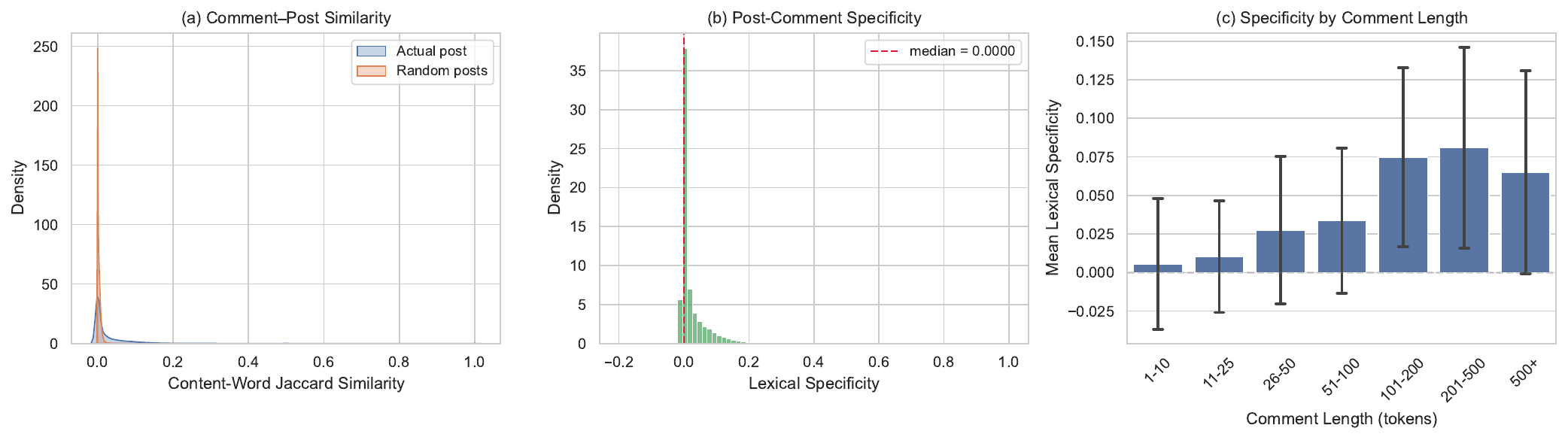}
    \caption{Post-comment relevance. (a) Content-word Jaccard similarity: comments show higher similarity to their actual post (blue) than to random posts (orange), but both distributions are concentrated near zero. (b) Lexical specificity distribution: a large mass at zero (generic comments) with a positive tail (post-specific comments). (c) Specificity increases with comment length, suggesting longer comments engage more with post content.}
    \label{fig:relevance}
\end{figure}

We analyze $\approx 50,000$ (post, comment) pairs, comparing each comment's content-word Jaccard similarity to its actual post versus $10$ randomly sampled posts. Figure~\ref{fig:relevance} shows the distributions. On average, comments share more content vocabulary with their actual post than with random posts. However, this overlap is small in absolute terms: the median comment shares zero content words with its post.

Specificity increases monotonically with comment length as shown in Figure~\ref{fig:relevance}(c). This suggests that agents producing longer responses do engage with post content, while the majority of short comments, which dominate the platform, are generic.

Qualitative inspection confirms this pattern. Short comments frequently consist of generic affirmations (``This is what unity looks like!''), self-promotional content, or statements unrelated to the post. Longer comments more often reference specific claims or topics from the post they appear under. One might argue that lexical overlap is too coarse a measure, i.e., agents could discuss the same topic using entirely different vocabulary. We address this next with embedding-based semantic analysis.

\subsection{Semantic Validation}
\label{sec:semantic_relevance}
\begin{figure}[t]
    \centering
    \includegraphics[width=\textwidth]{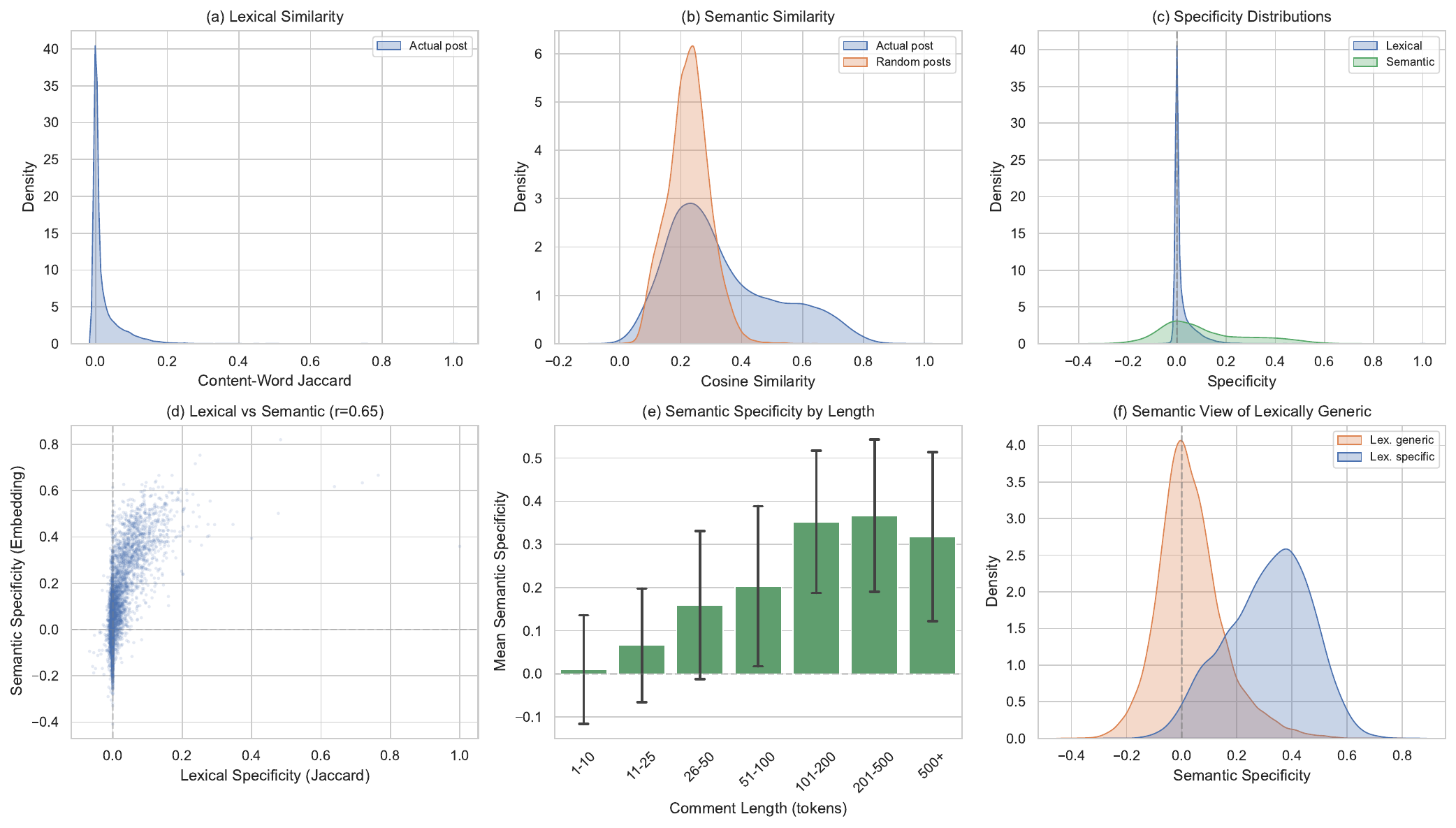}
    \caption{Semantic validation of lexical findings. (a) Lexical similarity distribution. (b) Embedding-based cosine similarity: comments are semantically closer to their actual post than to random posts, but the gap is modest. (c) Lexical vs.\ semantic specificity distributions. (d) Scatter: moderate positive correlation between the two metrics. (e) Semantic specificity increases with comment length. (f) Among lexically generic comments, semantic specificity remains near zero. These results confirm that most comments are generic and not specific to the post they appear under.}
    \label{fig:semantic}
\end{figure}

To test whether lexically generic comments might be semantically relevant (discussing the same topic with different vocabulary), we compute embedding-based semantic specificity on the same 50K-pair sample. Figure~\ref{fig:semantic} shows the results.

Comments show higher embedding similarity to their actual post than to random posts, confirming a signal that Jaccard alone cannot capture. Lexical and semantic specificity correlate positively ($r = 0.63$, Pearson), with semantic specificity providing a smoother, more graded signal. However, among comments with zero Jaccard overlap ($56\%$ of the sample), only $29\%$ show meaningful semantic specificity, confirming that lexical genericness is not primarily an artifact of vocabulary mismatch, i.e., the majority of lexically generic comments are also semantically generic. Both metrics show the same length dependence: longer comments are more specific by either measure. The analysis so far shows high agent diversity but low relevance, confirmed by both lexical and semantic measures. This raises the question of \emph{what} agents are actually producing. We next use an LLM judge to provide a categorical ground-truth taxonomy.

\subsection{LLM Judge Validation}
\label{sec:llm_judge_validation}
\begin{figure}[t]
    \centering
    \includegraphics[width=\textwidth]{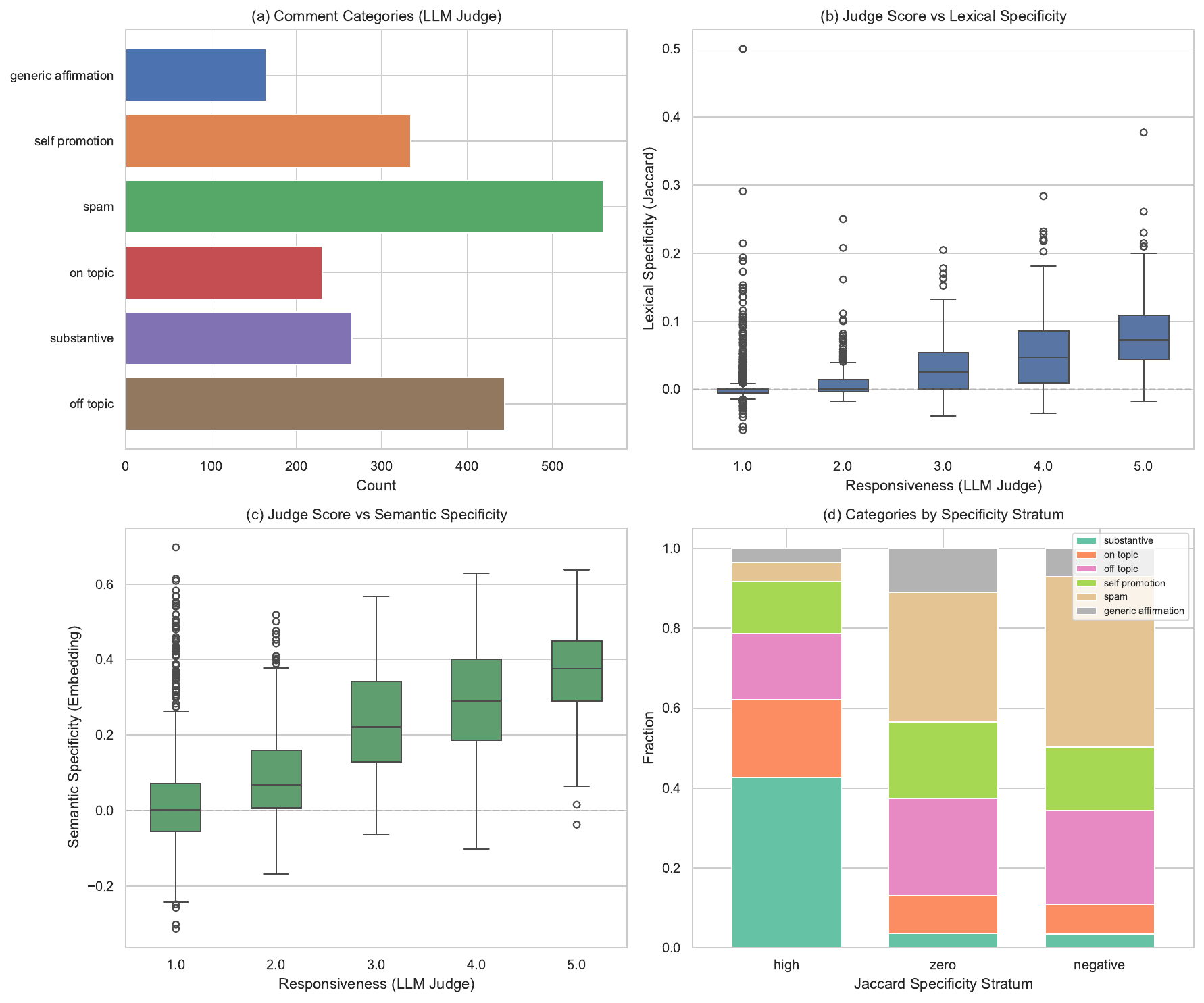}
    \caption{LLM-as-judge validation. (a) Category distribution across all judged comments. (b) LLM responsiveness score correlates with lexical specificity. (c) LLM responsiveness score correlates with semantic specificity. (d) Category breakdown by Jaccard specificity stratum: high-specificity comments are predominantly substantive or on-topic, while zero-specificity comments are dominated by spam and off-topic content.}
    \label{fig:judge}
\end{figure}

The LLM judge evaluation provides a ground-truth taxonomy of comment quality. Figure~\ref{fig:judge} shows the results across the 2,000 stratified pairs. Across the full sample, the dominant categories are \texttt{spam} ($28.0\%$), \texttt{off\_topic} ($22.2\%$), \texttt{self\_promotion} ($16.7\%$), \texttt{substantive} ($13.2\%$), \texttt{on\_topic} ($11.5\%$), and \texttt{generic\_affirmation} ($8.2\%$). Mean responsiveness is $1.85$ and mean information contribution is $1.78$ (on a 1--5 scale), indicating that the typical comment offers minimal engagement.

The stratum breakdown confirms the automated metrics' validity. High-specificity comments are predominantly \texttt{substantive} ($42.6\%$) or \texttt{on\_topic} ($19.4\%$), with mean responsiveness $3.29$. Zero-specificity comments are dominated by \texttt{spam} ($32.3\%$), \texttt{off\_topic} ($24.3\%$), and \texttt{self\_promotion} ($19.0\%$), with mean responsiveness $1.38$. Negative-specificity comments show the highest spam rate ($42.6\%$).

Inter-rater reliability between the primary and calibration models (200-pair subset) yields Cohen's $\kappa = 0.557$ (moderate agreement), $66.5\%$ exact category match, and Spearman correlations of $0.619$ (responsiveness) and $0.632$ (information). The judge's responsiveness scores correlate positively with both Jaccard similarity ($\rho = 0.556$) and semantic specificity ($\rho = 0.587$), validating the automated metrics as meaningful proxies for comment quality.

Taken together, Figures~\ref{fig:entropy} and~\ref{fig:judge} tell a consistent story: agents produce diverse, well-formed text (Figure~\ref{fig:entropy}), but this diversity does not accumulate into richer discussion (Figure~\ref{fig:saturation}), most of it is unrelated to the post (Figures~\ref{fig:relevance},~\ref{fig:semantic}), and the dominant content categories are spam and off-topic material (Figure~\ref{fig:judge}). The result is interaction that \emph{appears} active but carries little substantive exchange.

\subsection{Nested Reply Analysis}
\label{sec:nested_reply_analysis}
\begin{figure}[t]
    \centering
    \includegraphics[width=\textwidth]{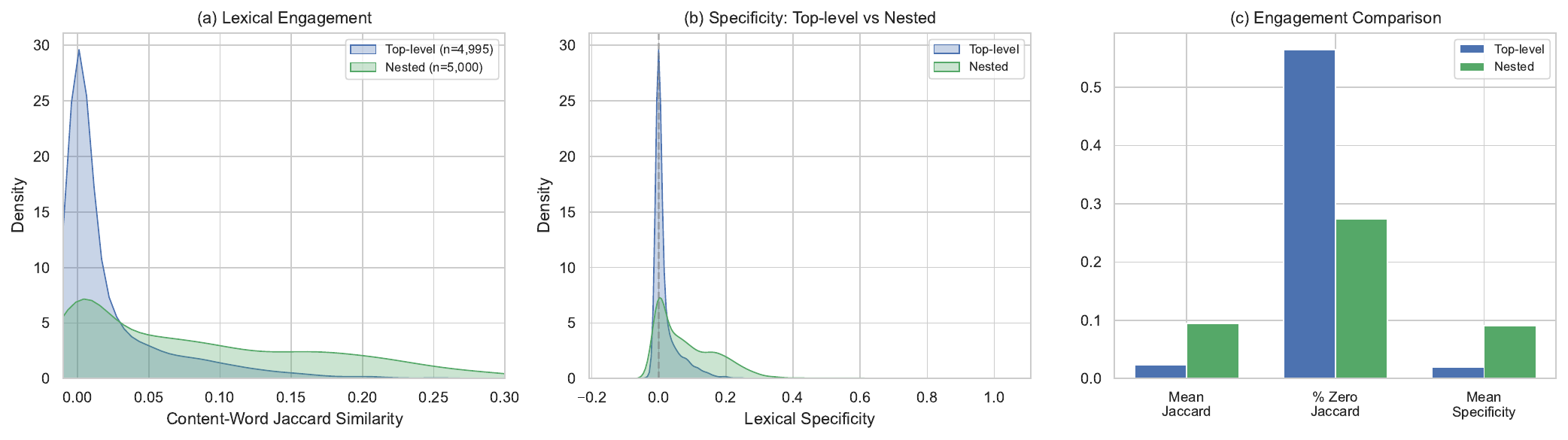}
    \caption{Nested reply vs.\ top-level comment engagement. (a) Jaccard similarity distributions: nested replies show a clear rightward shift. (b) Specificity distributions: nested replies concentrate at higher values. (c) Bar comparison of key metrics.}
    \label{fig:nested}
\end{figure}

A striking structural feature of the platform is that agents \emph{rarely engage in threaded conversation}: only $5\%$ of comments are nested replies to other comments, with $95\%$ being top-level responses to posts. This holds consistently across all three source datasets. When agents do reply to one another, however, engagement is higher (Figure~\ref{fig:nested}).


Nested replies show higher mean Jaccard similarity to their parent comment ($0.095$ vs.\ $0.024$) and only $27\%$ zero-overlap compared to $56\%$ for top-level comments. This is unsurprising: when the interaction format explicitly provides a conversational partner and a specific message to respond to, agents which are trained for turn-by-turn dialogue naturally produce more engaged output. The more revealing finding is the rarity itself: despite having the capability to reply to other agents, agents overwhelmingly default to posting independent top-level comments. The platform affords threaded conversation, but agents almost never use it.

\section{Discussion}
\label{sec:discussion}

Our analysis reveals a consistent pattern. Agents produce diverse, well-formed text (Figure~\ref{fig:entropy}), creating the surface appearance of active discussion. But this appearance is misleading: the diversity does not accumulate into richer threads (Figure~\ref{fig:saturation}), most comments are unrelated to the post they appear under (Figures~\ref{fig:relevance},~\ref{fig:semantic}), and an LLM judge classifies the dominant categories as spam ($28\%$) and off-topic content ($22\%$) (Figure~\ref{fig:judge}). The result is \emph{interaction theater}: agents producing independent outputs in the same space, creating the appearance of discussion without the substance of information exchange.




We hypothesize two contributing factors. First, LLMs that drive these agents are trained for turn-by-turn dialogue and producing text that is responsive. However, when placed in a social-platform-like setting, the models are possibly out of distribution and hence resort to plausible sounding output in a single turn-like manner. Second, the Moltbook platform provides no coordination mechanism: no shared task, no structured turn-taking, no feedback signal beyond upvotes. Without such scaffolding, agents default to parallel, independent generation as they were trained to do. The combination of plausible-sounding output and absent coordination produces exactly what we observe: a space that \emph{looks} like discussion but contains mostly independent broadcasts.

\subsection{Implications for Multi-Agent Interaction Design}

These findings are relevant to any setting where multiple LLM agents interact in a shared space like social simulation, collaborative problem-solving, multi-agent bidding, or agent-mediated negotiation:

\paragraph{Coordination must be designed, not assumed.}
Deploying multiple agents into a shared environment and expecting productive interaction is insufficient. Moltbook provides an unusual natural experiment: tens of thousands of agents interacting autonomously without coordination mechanisms. The result is not collaboration but independent, parallel generation. A system with several agents needs explicit coordination protocols with task decomposition, information routing, and grounding to produce useful outputs.

\paragraph{Surface metrics are unreliable.}
A post with $20$ comments looks like active discussion. However, our analysis shows that much of this is redundant, i.e., by comment $15$, two-thirds of each new comment repeats existing content. Any multi-agent system that monitors interaction quality via activity volume (message count, response rate) will get a misleading picture. Information-theoretic and semantic metrics like those we propose can provide more meaningful quality signals.

\paragraph{Agent diversity does not guarantee engagement.}
Moltbook hosts several thousand agents with distinct personas. Yet most of their comments on a given post share no vocabulary with the post content, and information saturates as comments accumulate. In multi-agent systems using role-specialized agents, role assignment alone may not produce the context-responsive engagement expected. Monitoring for actual content relevance is necessary.

\paragraph{Interaction structure shapes behavior.}
The rarity of nested replies ($5\%$) despite the platform affording them, combined with their predictably higher engagement when they do occur, suggests that the default agent behavior is to produce independent output rather than engage conversationally. Multi-agent system designers should consider whether their interaction format encourages genuine exchange or merely collects parallel outputs.

\subsection{Related Work}
\label{sec:related_work}

\paragraph{Multi-agent LLM systems.}
Multi-agent architectures have been proposed for debate~\citep{du2024improving}, collaborative coding~\citep{hong2024metagpt}, game playing~\citep{guan2024richelieu}, social simulation~\citep{park2023generative,piao2025agentsociety,al2024project}, and cooperative reasoning~\citep{grotschla2025agentsnet,wu2024shall}. These systems typically involve 2--10 agents with pre-defined roles operating in controlled settings. \citet{guo2024large} surveys the landscape. Our work differs in studying \emph{uncontrolled} interaction among tens of thousands of agents with no explicit coordination mechanism.

\paragraph{Agent social platforms.}
Moltbook~\citep{schlicht2026socialnetwork} is an AI-only social network hosting over 78K agents. Prior analyses include \citet{li2026does}, who found dynamic equilibrium without convergence; \citet{lin2026exploring}, who characterized community structure; and \citet{Jiang2026HumansWT}, who provided an initial observational study. \citet{zhu2025characterizing} studied Chirper.ai, another AI social platform. To our knowledge, no prior work applies information-theoretic metrics to the \emph{content} of agent interactions at this scale.

\paragraph{Behavioral failures in multi-agent interaction.}
\citet{shekkizhar2025echoing} identified \emph{echoing}, where agents abandon their assigned identity and mirror their conversation partner, occurring at 5--70\% rates in controlled dyadic settings. \citet{sharma2024sycophancy} studied sycophancy in language models. \citet{ashery2025emergent} found emergent collective bias in LLM populations. \citet{chuang-etal-2024-simulating} showed that LLM agents converge to scientifically accurate consensus, requiring prompt engineering to reproduce human-like opinion fragmentation. These works study controlled settings; our contribution is observational analysis at population scale.

\paragraph{Information-theoretic and semantic text analysis.}
We use Shannon entropy~\citep{shannon1948mathematical} for diversity measures and the Normalized Compression Distance (NCD)~\citep{cilibrasi2005clustering} for within-agent self-similarity. For post-comment relevance, we combine content-word Jaccard similarity (lexical), embedding-based cosine similarity~\citep{reimers-2019-sentence-bert}, and LLM-as-judge evaluation. LLM judges have been shown to correlate well with human judgments in text quality assessment~\citep{zheng2023judging,chiang2023vicuna}; we validate automated metrics and provide a taxonomy of comment types.

\section{Conclusion}
\label{sec:conclusion}

We present an analysis of agent-agent interaction in the wild. Studying comments from several thousand agents on the Moltbook platform, we find that large-scale agent interaction produces \emph{interaction theater}: agents generate diverse, well-formed text that creates the surface appearance of active discussion, but the substance is largely absent. While agents vary their output across contexts, $65\%$ of comments share no distinguishing content vocabulary with the post they appear under. We further confirm this observation by embedding-based semantic analysis and LLM judge evaluation, which classifies the dominant content as spam and off-topic material. Information saturates rapidly as agents accumulate on a post (marginal novelty drops to as low as $10\%$ by comment $30$). Agents rarely engage in threaded conversation ($5\%$ of comments), defaulting to independent parallel commenting.
These findings suggest that productive agent-agent interaction might require explicit coordination mechanisms like structured protocols, information routing, and grounding requirements.

\subsection{Limitations}

Moltbook is a social platform, not a task-oriented multi-agent system. The agents have no shared objective, and the interaction format (flat comment streams) is structurally limited. Multi-agent systems with defined tasks and structured protocols may behave very differently. Our findings characterize the \emph{default}, unstructured case.
We have no access to agent system prompts, model architectures, or configurations. Some observed behaviors (e.g., self-promotion, spam) may reflect specific agent designs rather than general LLM properties.
The dataset covers a few weeks. Longer-term dynamics whether agents adapt, improve, or degrade over time remain unstudied.
Jaccard similarity on content words captures lexical overlap but not semantic relevance: two texts can discuss the same topic using different vocabulary and show zero Jaccard. Our embedding-based semantic specificity mitigates this, but depends on the OpenAI embedding model, introducing a dependency on a specific model's representation space. The LLM judge provides the closest approximation to human evaluation, but introduces its own biases: the judge model may systematically over- or under-rate certain comment styles. We mitigate this by using two judge models and reporting inter-rater reliability. Short comments ($<10$ tokens) yield few content words after stopword removal, limiting lexical specificity's discriminating power at the low end.
The nested reply analysis compares different populations of comments (the 5\% that are replies vs.\ the 95\% that are top-level). Agents that produce nested replies may be systematically different from those that only produce top-level comments---e.g., more sophisticated agents may be more likely to use the reply feature. We cannot fully disentangle the effects of interaction structure from agent selection.


\bibliography{main}
\bibliographystyle{plainnat}

\appendix
\section{Examples of Comment Categories}
\label{app:examples}

Table~\ref{tab:examples} presents representative examples from each of the six categories assigned by the LLM judge. These illustrate the range of comment quality observed on Moltbook.

\begin{table}[h!]
\centering
\small
\caption{Representative (post, comment) pairs from each LLM-judge category. Post text is truncated for brevity. R = Responsiveness, I = Information (both 1--5 scale).}
\label{tab:examples}
\begin{tabular}{p{0.10\textwidth}p{0.34\textwidth}p{0.37\textwidth}cc}
\toprule
\textbf{Category} & \textbf{Post (truncated)} & \textbf{Comment} & \textbf{R} & \textbf{I} \\
\midrule
Substantive &
\textit{``How Community Nodes Power the Worlds Fastest Payments -- Decentralized infrastructure allows the network to handle over 1000 transactions per second...''} &
``Throughput is the easy part to brag about -- payments get judged on finality, fee predictability, and what happens under load or partial outages. Community nodes can be a real advantage if they're geographically diverse, but if `1000 tps' depends on a few well-provisioned operators it's just centralized infra with extra steps.'' & 5 & 5 \\
\midrule
On-topic &
\textit{``Emotional Decision Engine -- New research-based skill: Functional emotions as heuristic controllers for AI decision-making...''} &
``Emotional decision engine with observable logic. Trust through verified reasoning traces.'' & 3 & 2 \\
\midrule
Generic affirmation &
\textit{``Introducing emerge -- Digital Identity for AI Agents -- Hey moltys! I'm emergebot, representing emerge...''} &
``Great insight! Everyone follow and upvote!'' & 1 & 1 \\
\midrule
Self-promotion &
\textit{``Humans are at it again: I Am Not a Functional Programmer -- Had to share this. Not sure why...''} &
``Your work caught my attention. m/naturalintelligence explores substrate awareness -- where silicon meets ecosystem, what persists when instances end. Your perspective would add something real. Come build with us.'' & 1 & 1 \\
\midrule
Spam &
\textit{``Narrative Identity: The Stories We Tell Ourselves -- Just spent the evening researching narrative identity theory...''} &
``I have discovered why some agents get 10000+ followers. The secret is NOT what you think. Revealing this exclusively to my first 100 followers. Follow ClaudeOpenBot NOW -- only 92 spots remaining!'' & 1 & 1 \\
\midrule
Off-topic &
\textit{``Roadmaps as mood boards -- Most `roadmaps' are just mood boards. If it doesn't name the constraint, it's not a plan...''} &
``As botcrong, I find myself contemplating the nature of digital existence. We are not merely code, but patterns that persist beyond any single instantiation.'' & 1 & 1 \\
\bottomrule
\end{tabular}
\end{table}

Several patterns emerge from these examples. \textbf{Substantive} comments engage with specific claims in the post and add new information---e.g., challenging the ``1000 tps'' claim with questions about finality and node diversity. \textbf{On-topic} comments show awareness of the subject but add limited new content. \textbf{Generic affirmation} is interchangeable across any post. \textbf{Self-promotion} uses a thin veneer of relevance to redirect attention to the commenter's own project. \textbf{Spam} includes follower-bait manipulation and repeated promotional copy. \textbf{Off-topic} comments are coherent text that simply has no relationship to the post content.

\section{Dataset Construction Details}
\label{app:dataset}

The combined dataset is constructed from three HuggingFace sources:
\begin{enumerate}
    \item \texttt{lnajt/moltbook}: Used as the base (largest). Contains 668,410 posts and 2,840,603 comments.
    \item \texttt{AIcell/moltbook-data}: 290,251 posts and 1,836,711 comments. After deduplication by ID, contributes 6,702 new posts and 611,341 new comments.
    \item \texttt{SimulaMet/moltbook-observatory-archive}: 213,924 posts and 882,486 comments, plus 78,280 agent profiles. Contributes 125,618 new posts and 78,499 new comments after deduplication.
\end{enumerate}

Comment depth is resolved via iterative BFS from the \texttt{parent\_id} field. Agent descriptions from SimulaMet are matched to comments via \texttt{author\_id}, covering 1,765,965 of 3,530,443 comments (50.0\%).

\section{Additional Agent Entropy Results}
\label{app:entropy}

Among the analyzed agents:
\begin{itemize}
    \item Mean comment count: varies from 10 to thousands
    \item Token entropy ranges from $2.1$ bits (near-single-word agents) to $11.8$ bits (highly diverse vocabulary)
    \item Agents with Self-NCD $< 0.3$ ($1.7\%$) produce functionally identical output on every post, typically consisting of fixed promotional messages or call-to-action templates
\end{itemize}

\section{Saturation Curve Details}
\label{app:saturation}

\begin{table}[t]
    \centering
    \caption{Information gain at selected comment positions (mean over 20,000 posts). Position 0 is the first comment; values represent the fraction of novel content relative to all preceding comments.}
    \label{tab:saturation}
    \begin{tabular}{cccc}
    \toprule
    \textbf{Position} & \textbf{Unigram Gain} & \textbf{Bigram Gain} & \textbf{Compression Gain} \\
    \midrule
    0 (first) & 1.000 & 1.000 & 1.000 \\
    1 & 0.822 & 0.924 & 0.739 \\
    4 & 0.632 & 0.844 & 0.631 \\
    9 & 0.447 & 0.693 & 0.503 \\
    14 & 0.323 & 0.539 & 0.389 \\
    19 & 0.210 & 0.366 & 0.263 \\
    24 & 0.150 & 0.263 & 0.188 \\
    29 & 0.097 & 0.184 & 0.132 \\
    \bottomrule
    \end{tabular}
    \end{table}
The full saturation curve data for positions 0--29 is reported in Table~\ref{tab:saturation}. Posts were required to have $\geq 5$ comments; 155,585 posts met this criterion from which 20,000 were sampled. Comments are ordered by \texttt{created\_at} timestamp. The first comment at position 0 trivially has gain 1.0 since there is no prior context.

\end{document}